# Review of Deep Reinforcement Learning for Autonomous Driving

B. Udugama, Middlesex University, M00734040

*Abstract*—Since deep neural networks' resurgence, reinforcement learning has gradually strengthened and surpassed humans in many conventional games. However, it is not easy to copy these accomplishments to autonomous driving because state spaces are immensely complicated in the real world and action spaces are continuous and fine control is necessary. Besides, autonomous driving systems must also maintain their functionality regardless of the environment's complexity. The deep reinforcement learning domain (DRL) has become a robust learning framework to handle complex policies in high dimensional surroundings with deep representation learning. This research outlines deep, reinforcement learning algorithms (DRL). It presents a nomenclature of autonomous driving in which DRL techniques have been used, thus discussing important computational issues in evaluating autonomous driving agents in the real environment. Instead, it involves similar but not standard RL techniques, adjoining fields such as emulation of actions, modelling imitation, inverse reinforcement learning. The simulators' role in training agents is addressed, as are the methods for validating, checking and robustness of existing RL solutions.

## I. INTRODUCTION

FOR a decade, the autonomous car has been in the news and continues to dominate auto headlines. Researchers, robotics organizations and the automotive industry have been fascinated by an autonomous vehicle. Human driving is accident-prone. The failure of humans to obtain smarter spontaneous driving decisions triggers road collisions, asset loss and fatalities[1]. The autonomous vehicle offers us the capability to restore an error-prone human driver by offering reassurance and protection. Driverless systems consist of various functions at the perception level that has now attained high accuracy due to deep learning architectures. In addition to perception, DRL autonomous driving technologies have addressed several challenges in which conventional supervised learning techniques are no longer valid. First, as the estimation of the agent's behaviour alters upcoming sensor information gained from the context where the autonomous agent works, the role of optimum driving speed in a metropolitan setting, for example, adjusts. Second, the regulatory factors such as the time of collision, the longitudinal deviation w.r.t to the optimum route of the autonomous system reflect both the dynamics of the agent and environmental ambiguity[2]. Of this kind, challenges will entail the concept of the stochastic cost function to be maximized. This describes a higher feature space provided with

Specific settings wherein the agent & ecosystem has been studied, which is significant. In these kinds of situations, researchers attempt to overcome a systematic decision-making framework formulated under the classic Reinforcement Learning (RL) conditions, where the system is expected to observe and perceive the ecosystem and, therefore, behave adequately at every moment. Optimum behaviour is attributed to the policy[3].

The principles of reinforcement learning, the classification of tasks where RL is a probable approach, particularly in cruising strategy, predictive cognition, trajectory and navigation planning, and low-level control system architecture, are discussed in this survey. This analysis also reflects on RL's numerous engagements in the context of autonomous driving (AD). Ultimately, discuss deploying modern RL techniques like imitation learning and deep Q learning by showing the main constraints and consequences[1].

The main aspects of this review:
- Self-contained RL overview for the automotive sector.
- Comprehensive literature overview about the use of RL for various automated driving assignments.
- Analysis of the main problems and prospects for RL applying to automated vehicles in the real environment.

## II. CONSTITUENTS OF AUTONOMOUS DRIVING SYSTEM

Fig. 1. contains the specific parts of an AD unit's Motion planning, showing the flow from the route planning to the self-control actuation. The sensor design involves several sets of sensors, radars and LIDARs in a typical autonomous driving vehicle and a GPS-GNSS system for accurate positioning and inertial measurement units that offer 3D localization to the device[15].

The purpose of the perception component is to produce an intermediary level description of the system's status that is then used by a policymaking technique that will effectively establish the operational policy. This primary condition would consist of lane placement, drivable region, symbolic location such as pedestrians and vehicles, state of others, and traffic lights. Perception problems spread to the remainder of the communication chain[6]. Robust technology realization is essential for safety; therefore, redundant options improve confidence in detection. This is accomplished by combining multiple vision tasks, including semantic segmentation, motion

M00734040 Bavantha Udugama

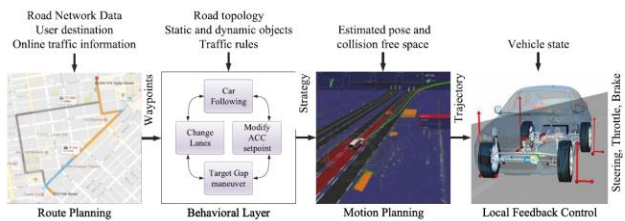

Fig. 1. Layers of motion planning for AD systems[5]

estimation, estimation of depth, identification of soiling, which is typically easily unified directly through a multi-task design.

### A. Understand the Surrounding

The abstract mid-level representation of the perception state from the perception module is mapped by this main module on the higher-level intervention or even decision-making module. Abstractly, this specific portion groups three tasks: Comprehension of the scene, decision and even preparation. As seen in figure one module, it is assembled on top of the algorithmic localization or detection tasks to establish a higher-level understanding of the scene. It attempts to vigorously simplify situations by fusing heterogeneous sensor capital as the information becomes even more abstract[4].

This merger material offers a broad and condensed context for the components of the decision. Fusion provides a sceptical sensor image of the eco system and models the sensor noise and even uncertainties of identification across many modalities such as LIDAR, radar, video, ultrasound. This essentially involves weighting the projections by using a process based on values.

### B. Localization and Mapping

One of the crucial foundations of autonomous driving is visualization. When an area is surveyed, it is easy to find the vehicle's actual location on the map. The first coherent AD presentations relied largely on localization to pre-mapped areas. Conventional mapping techniques are improved by semantic object recognition for coherent disambiguation because of the extent of the query. In particular, localized high-definition maps can be seen as a precedent for object detection.

### C. Route Planning and policy

In the AD pipeline, route preparation is a key factor. This module is required to create motion-level controls that manoeuvre the car, providing a route-level plan from HD maps or GPS based maps.

### D. Controlling the Autonomous system

A controller determines the speed, steering angle and decelerating behaviour expected by a pre-established map such as Google maps over each point in the road or appropriate driving recording of the same values at each waypoint. Path following, by contrast, includes a terrestrial model of the automobile's dynamics viewing the waypoints over a given duration in sequence[7].

## III. RL FOR AUTONOMOUS DRIVING TASKS

AD tasks where RL could be implemented include optimization of controllers, scheduling of paths and optimization of trajectories, movement planning and dynamic path planning, expansion of high-level driving policies for complex navigation tasks, outcome-based policy learning for expressways, crossings, mergers and splits, reward learning with converse reinforcement learning from intelligence expert data. Then briefly study the state space, action space and rewards mechanisms in these ecosystems before exploring the DRL frameworks for AD tasks[6].

Implementing adequate state spaces, action spaces, and rewards mechanisms is essential in order to effectively apply DRL to automated driving assignments. Frequently utilized state-space characteristics for automated driving include: ego-vehicle location, heading, and velocity, as well as other constraints in the ego-vehicle sensor view spectrum [5]. This is further improved by lane details like lane number, route contour, ego-vehicle context and projected trajectory, longitudinal data such as time to collision, and ultimately scenario relevant data such as traffic regulations and locations of the signal( see Fig. 2.).

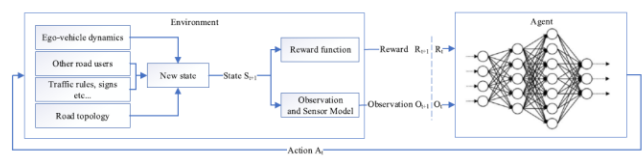

Fig. 2. deep reinforcement learning based autonomous driving[4]

## IV. REINFORCEMENT LEARNING – MODELING

### A. Modelling the Autonomous system

A key aspect of the learning experience is modelling the ego-vehicle movement as it poses the tradeoff question between model accuracy and computational capital. Since RL strategies use a large number of episodes to evaluate optimum strategy, the environmental phase time, which strongly depends on the vehicle dynamics model's assessment time, has a profound effect on training time. More complex models with a larger number of parameters and complicated tyre models must be chosen from the simplest kinematic model to more advanced dynamics models[15].

Special simulators are also used to model traffic and surrounding vehicles. Using cellular automation models, some authors build their environments. Some use MOBIL, which is a general model (minimizing lane shift-induced overall braking) to extract discretionary and obligatory lane change laws for a broad class of car-following models; the Intelligent Driving Model (IDM), a single-lane continuous microscopic model[3].

### B. Simulation

To gain complete control over the model, some writers build self-made environments, although there are commercial and open-source environments that can include this functionality.

TABLE I
SIMULATORS FOR RL APPLICATIONS IN ADVANCED DRIVING ASSISTANCE SYSTEMS (ADAS) AND AUTONOMOUS DRIVING.

| Simulator | Description |
|---|---|
| CARLA [8] | Urban simulator, Camera & LIDAR streams, with depth & semantic segmentation, Location information |
| TORCS [9] | Racing Simulator, Camera stream, agent positions, testing control policies for vehicles |
| AIRSIM [10] | Camera stream with depth and semantic segmentation, support for drones |
| GAZEBO (ROS) [11] | Multi-robot physics simulator employed for path planning & vehicle control in complex 2D & 3D maps |
| SUMO [12] | Macro-scale modelling of traffic in cities motion planning simulators are used |
| DeepDrive [13] | Driving simulator based on unreal, providing multi-camera (eight) stream with depth |
| Constellation [14] | NVIDIA DRIVE ConstellationTM simulates camera, LIDAR & radar for AD (Proprietary) |

Any of them used in recent research into motion planning with RL are briefly described in Table 1.

*C. Actions Space*

The choice of action configuration depends highly on the vehicle model and task configured for the reinforcement learning problem. Although it is possible to see two key layers of control: one is the basic control of the car by regulating de accelerating and accelerating orders, and the other operates on the behavioural layer and determines strategic level decisions, such as lane shift, lane management, Accurate reference point setting, etc. The agent gives an order at this stage to low-level controllers who determine the real trajectory. Just a few papers deal with the layer of motion planning, where the mission specifies the endpoints [11]. In comparison, few papers deviate from constraints on vehicle motion and produce behaviour by moving onto a grid, such as in classic microscopic models of cellular automatics[3].

*D. Rewarding Functions*

The agent attempts to fulfil a mission during preparation, normally containing of more than one move. An episode is called this mission. An episode ends until one of the following criteria is met:

- The agent executes the role efficiently.
- A previously specified stage is reached by the episode
- A terminating status enhances.

The first two cases are insignificant and rely on the real problem's nature. Terminal cases are usually circumstances in which the agent enters a position from which it is difficult to perform the actual mission, or the agent commits an intolerable error. Vehicle motion preparation agents normally use termination circumstances, such as accident or exiting the track or lane with other members or barriers, since the episode eventually concludes with these two. There are lighter ways, with examples of having too high a tangent angle to the track or reaching too close to other people, where the episode terminates with failure before the crash occurs. These "before crash" terminations accelerate the training by taking the loss details forward in time, while caution is required in their design[15].

The first significant factor is the pacing of the incentive, where the builder of the reinforcement learning approach has to select a combination of both the pros and cons of the following strategies:

- Rewarding and discounting it back, which could result in a slower learning process while reducing the policy's human-driven shaping.
- Naturally, the discount often occurs in this approach, providing immediate reward at each stage by measuring the current situation, resulting in considerably faster learning, but the choice of the immediate reward strongly impacts the developed strategy, which often escapes the strategy.
- In predefined times or travel distances[6], or where a positive or poor decision takes place, an intermediate option might be to offer an incentive.

*E. Observation Space*

The room for perception explains the universe to the agent. It needs to have adequate information to choose the required action, so it includes - based on the mission - the following knowledge:

*1) Vehicle State Observation*

The most widely used and often the easiest observation for the ego vehicle consists of the unceasing variables of ($|e|$, $v$, $\Theta_e$) representing the lateral direction from the centre-line of the lane, vehicle speed, and yaw angle correspondingly for lane holding, navigation, easy racing, overtaking, or manoeuvring activities. (see Fig. 3).

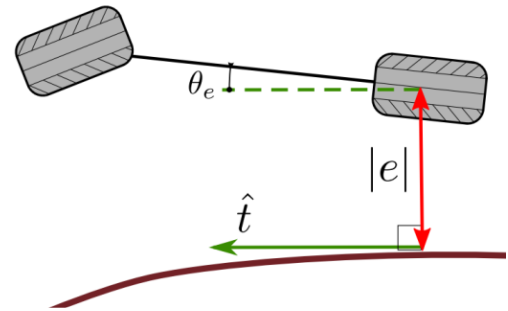

Fig.3. Basic vehicle state model[1]

*2) Environment Observation*

Having knowledge about the vehicle world and representing it to the learning agent reflects a high degree of diversity in the literature. It is possible to observe different degrees of sensor abstractions:

- Perception level, where camera images, lidar or radar data are transferred to the agent
- The intermediate stage, where idealized knowledge about sensors is provided
- Ground truth stage, where all information that is measurable and not detectable is given.

The structure of the sensor model also affects the Deep RL agent's neural network structure since image-like or array-like inputs infer 2D or 1D CNN structures, whereas a single dense network results in a simple collection of scalar information. There are examples of combining these two

TABLE II
AUTONOMOUS DRIVING TASKS WITH REUQIRED DRL OT LEARN POLICY BEHAVIOUR

| AD Task | DRL METHOD & DESCRIPTION | IMPROVEMENTS & TRADEOFFS |
|---|---|---|
| Lane Keep | A DRL method for discrete actions (DQN) and continuous actions (DDAC) using the TORCS simulator is proposed by the authors[2]. Authors[3] are studying discrete and continuous policies to follow the lane and optimise average velocity using DQNs and Deep Deterministic Actor Critic (DDAC). | 1. This research concludes that the use of continuous behaviour creates smoother pathways, although it leads to more restrictive termination conditions on the negative side and longer convergence time to understand. 2. For quicker integration & improved efficiency, eliminating memory replay in DQNs helps. The one hot action space encoding resulted in sudden power of steering. Although the continuing strategy of DDAC helps smooth the acts and delivers improved results. |
| Lane Change | Authors[15] are using Q-learning to learn a no-operation guideline for egovehicle, lane shift to left/right, accelerate/decelerate. | Compared to conventional approaches, this approach is more stable, and consists of identifying fixed way points, velocity profiles, and direction curvature to be taken by the ego car. |
| Ramp Merging | Authors[5] suggest recurrent architectures, namely LSTMs, to model long-term dependencies for the merger into a highway ramp of ego automobiles. | To execute the merging more robustly, historical experience of state knowledge is used. |
| Intersections | To negotiate intersection, authors use DQN to test the Q-value for state-action pairs[15],, | Author-defined Creep-Go behaviour allow the vehicle to more securely navigate intersections with small spaces and visibility. |
| Motion Planning | An improved A¤ algorithm is proposed by the authors[88] to learn a heuristic function using deep neural networks over image-based obstacle map input. | Smooth vehicle control behaviour and increased performance compared to multi-step DQNN |

kinds of inputs. The network thus has to have two distinct types of input layers[3].

## V. EVENT-BASED CLASSIFICATION OF THE APPROACHES

While machine learning may be assumed to provide an overall end-to-end approach to autonomous driving, the review of recent literature indicates that research on Reinforcement Learning may provide answers to some sub-tasks of this problem. The articles can be structured around these problems in recent years, where a well-dedicated condition or case is selected and investigated whether it can be overcome by a self-learning agent[5].

### A. Following a car

The simplest challenge in this survey is to follow vehicles, where the question is articulated as follows: There are two simulation participants, a leading vehicle and the following vehicle, each retaining their side positions in a lane, and the following vehicle changes its longitudinal velocity to ensure a safe subsequent distance. The space out of observation consists of the tuple (v, dv, ds), representing agent velocity, lead velocity difference, and distance of headway[4].

### B. Lane following

Lane-keeping or following the trajectory is still a basic control task, but this concern focuses on lateral control, as opposed to car follow-up. There are two distinct approaches to the observation room in these studies: One is the lateral direction and angle of the vehicle in the road, "ground truth," while the second is a front camera view. Naturally, the agents use external simulators, TORCS, and GAZEBO/ROS in these instances for image-based control. The gap from the centerline of the lane is almost often regarded by incentive programmes as an instant reward. It is important to remember that these agents barely consider the dynamics of the vehicle and, oddly, do not rely on collective longitudinal regulation[15].

### C. Ramp Merging

The ramp merge dilemma deals with the highway on-ramp situation, where the ego vehicle has to locate the necessary distance to get on the highway between two vehicles. In the simplest method, the longitudinal regulation where the agent approaches this position is available for learning, as can be seen in—other papers, such as using complete power of steering and acceleration. The linear acceleration of the car accelerates and decelerates in the acts, and the ego vehicle keeps its lane when performing these actions. The "lane change left" and "lane change right" behaviour indicate lateral motion[2].

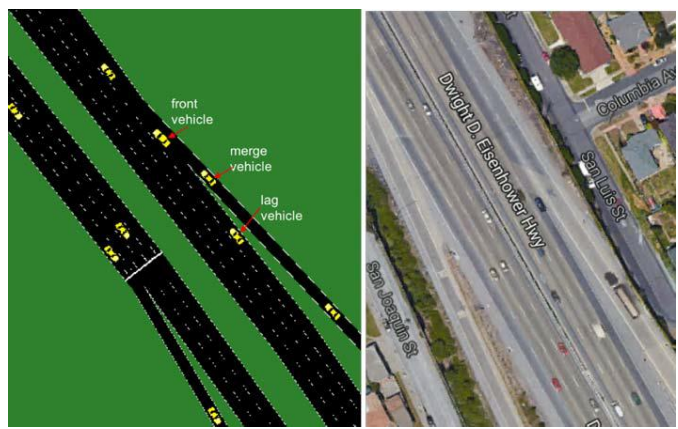

Fig.4. Ramp merge: (a) simulated scenario and (b) real-world location[3]

### D. Driving in Stream Of Traffic

In recent articles, the most complex situation discussed is where the autonomous agent is driving in traffic. Naturally, the topology of the network, the quantity and behaviour of the adjacent vehicles, the operation of traffic laws, and many other features also make this role scalable. In the previous pages, such as lane-keeping, or car trailing, sub-tasks of this scenario have been examined[8].

## VI. CONCLUSIONS

In real-world autonomous driving systems, reinforcement learning is still an active and emerging field. While a few commercial implementations are successful, relatively little literature or large-scale public databases are available. Therefore, we were inspired to formalize and coordinate RL autonomous driving implementations. Interacting agents are interested in autonomous driving situations that need

M00734040 Bavantha Udugama

negotiation and complex decision making that fits RL. However, in order to provide advanced ideas that we address in-depth, there are more problems to be overcome. Detailed theoretical reinforcement learning is discussed in this work.

Latest advances in the area have demonstrated that numerous deep reinforcement learning methods can be successfully used for various stages of motion planning problems for autonomous vehicles, but several questions remain unanswered. The key benefit of these approaches is that unstructured data such as raw or slightly pre-processed radar or camera-based image information can be treated.

The comparatively low computational criteria of the trained network are one of the key advantages of using deep neural networks trained by a reinforcement learning agent in motion planning. While this method requires a large number of trials in the learning phase to obtain adequate knowledge, as stated before, for basic problems of convex optimization, the mechanism converges easily. However, the preparation can rapidly hit millions of measures with complicated situations, meaning only one setup of hyperparameters or incentive hypothesis can last hours or even days.

Since complex reinforcement learning tasks involve ongoing iteration on the design of the environment, network configuration, incentive scheme, or even the algorithm itself, it is a time-consuming activity to design such a method. The measurement time depends heavily on the delegated computing resources and the required outcome interpretation and inference. On this basis, it is not shocking that most articles nowadays deal with small subtasks of motion planning, and the most complicated situations can not be found in the literature, such as travelling in urban traffic. RL itself, like many heuristics, has a tradeoff between efficiency and the need for capital.

The principal purpose of reinforcement learning is to statistically optimize the long-term incentive. Nevertheless, the main priority is the avoidance of injuries for vehicle control activities. Although the use of behaviour that triggers significant negative rewards does not inherently eliminate RL, other strategies must control the hazards. In several ways, the literature discusses protection and threats, for which[4] offers an exemplary overview. In this field, two principal directions can be separated. The approaches using the optimization criteria are included in one group of solutions, while the other group includes algorithms that change the discovery process. One has some choices for adjusting optimization parameters as well. The worst-case criterion is the first. Addressing the worst-case situations solves the concerns created by the uncertainty resulting from the stochastic instability of the system and the parameter uncertainties. The risk-sensitive criteria are the second choice. In this circumstance, a scalar parameter, a so-called risk susceptibility parameter, is applied to the loss function to control the degree of risk. Finally, it is possible to use a restricted Markov decision process (MDP), where the default MDP tuple is expanded with a constraint set that must be satisfied by the policy function.

Contrary to the classic exploration approach, changing the exploration phase is an alternative, which means that the agent knows something from scratch. That also leads to disastrous situations with vehicle control applications. In comparison, fully unintended discovery techniques spend a lot of time investigating the meaningless areas of the underlying state space, which is particularly important in broad and continuous state spaces. Two key directions are available. Through applying external intelligence, one guides the discovery process, while the other uses risk assessment. Through demonstrating the fascinating or dangerous sections of state space, the demonstrator may also lead the exploration online. And, ultimately, a supervisory control system will follow challenging constraints.

Overall, a dynamically changing field is the principle of stable RL. The subject's importance is unquestionable from the point of view of vehicle regulation, not only for safety but also for the reduction of the state and the room for intervention. The preference of troublesome, so-called corner cases from a large range of unrelated conditions is one of the major problems with preparation and validation.

In general, three paths to narrowing the gap exist:
- Identification of the system, aiming to adapt the simulation to reality.
- Domain adaptation helps to learn a well model from a source distribution of data on a separate target data distribution.
- Randomization of the domain targeted learning in a very randomized environment that covers the target and makes the agent resilient.

The tradeoff between the completely modelled system and feasibility was discussed, so identifying the system is not defined here. One aims to locate the transition strategy between the virtual and the actual representations during Domain adaptation. As an example, this transition can be solved by a semantically segmented image for image sequences taken from a front-facing camera. In [2], the two realms meet at the segmented stage in the centre, while in [1], the authors attempt to build "realistic" training images using generative adversarial networks (GAN) [7].

Overall, many problems in this area remain to be addressed, such as environmental detail and sensor simulation, computational specifications, transferability to actual systems, robustness, and agent validation. Due to these concerns, it can be claimed that reinforcement learning is not an adequate method for automotive motion planning. However, when combined with other approaches, it can be very useful in solving complex optimization challenges.